\documentclass[runningheads]{llncs}

\usepackage{graphicx}
\usepackage{subfigure}
\usepackage{amsmath}
\usepackage{amssymb}
\usepackage{cite}
\usepackage{multirow}
\usepackage{tabularx}

\begin{document}
\newcolumntype{C}[1]{>{\centering\arraybackslash}p{#1}}
\title{Pedestrian Trajectory Prediction with Structured Memory Hierarchies}
%
%

\author{Tharindu~Fernando \inst{1} \and
Simon Denman\inst{1} \and
Sridha Sridharan\inst{1} \and
Clinton Fookes\inst{1}}
\authorrunning{T. Fernando et al.}
%
\institute{Image and Video Research Laboratory, SAIVT Research Program, \\ Queensland University of Technology, Brisbane, Australia  \\
\email{\{t.warnakulasuriya, s.denman, s.sridharan, c.fookes\}@qut.edu.au}
}
\maketitle              
\begin{abstract}
This paper presents a novel framework for human trajectory prediction based on multimodal data (video and radar). Motivated by recent neuroscience discoveries, we propose incorporating a structured memory component in the human trajectory prediction pipeline to capture historical information to improve performance. We introduce structured LSTM cells for modelling the memory content hierarchically, preserving the spatiotemporal structure of the information and enabling us to capture both short-term and long-term context. We demonstrate how this architecture can be extended to integrate salient information from multiple modalities to automatically store and retrieve important information for decision making without any supervision. We evaluate the effectiveness of the proposed models on a novel multimodal dataset that we introduce, consisting of 40,000 pedestrian trajectories, acquired jointly from a radar system and a CCTV camera system installed in a public place. The performance is also evaluated on the publicly available New York Grand Central pedestrian database. In both settings, the proposed models demonstrate their capability to better anticipate future pedestrian motion compared to existing state of the art.

\end{abstract}

\section{Introduction}

Understanding and predicting crowd behaviour is an important topic due to its myriad applications (surveillance, event detection, traffic flow, etc). However this remains a challenging problem due to the complex nature of human behaviour and the lack of attention that researchers pay to human navigational patterns when developing machine learning models. 

Recent neuroscience studies have revealed that humans utilise map and grid like structures for navigation \cite{madl2016exploring,epstein2017cognitive}. The human brain builds a unified representation of the spatial environment, which is stored in the hippocampus \cite{fanselow2010dorsal} and guides the decision making process. 
Further studies \cite{brun2008progressive} provide strong evidence towards a hierarchical spatial representation of these maps. Additionally in \cite{derdikman2010manifold,gobet2001chunking} authors have observed multiple representations of structured maps instead of one single map in the long-term memory. This idea was explored in \cite{parisotto2018} using structured memory for Deep Reinforcement Learning. To generate an output at a particular time step, the system passes the memory content through a series of convolution layers to summarise the content. We argue this is inefficient and could lead to a loss of information when modelling large spatial areas. 

Motivated by recent neuroscience \cite{madl2016exploring,epstein2017cognitive} and deep reinforcement leaning \cite{parisotto2018} studies, we utilise a structured memory to predict human navigational behaviour. In particular such a memory structure allows a machine learning algorithm to exploit historical knowledge about the spatial structure of the environment, and reason and plan ahead, instead of generating reflexive behaviour based on the current context. Novel contributions of this paper are summarised as follows:


\begin{itemize}
\item We introduce a novel neural memory architecture which effectively captures the spatiotemporal structure of the environment.
\item  We propose structured LSTM (St-LSTM) cells, which model the structured memory hierarchically, preserving the memories' spatiotemporal structure. 
\item We incorporate the neural memory network into a human trajectory prediction pipeline where it learns to automatically store and retrieve important information for decision making without any supervision.
\item We introduce a novel multimodal dataset for human trajectory prediction, containing more than 40,000 trajectories from Radar and CCTV streams.  
\item We demonstrate how the semantic information from multiple input streams can be captured through multiple memory components and propose an effective fusion scheme that preserves the spatiotemporal structure. 
\item We provide extensive evaluations of the proposed method using multiple public benchmarks, where the proposed method is capable of imitating human navigation behaviour and outperforms state-of-the-art methods. 
\end{itemize}


\section{Related Work}
The related literature can be broadly categorised into human behaviour prediction approaches, introduced in Sec \ref{sec:hba}; neural memory architectures, presented in Sec. \ref{sec:nma}; and multimodal information fusion which we review in Sec. \ref{sec:mif}.
\subsection{Human Behaviour Prediction}
\label{sec:hba}
Before the dawn of deep learning, Social Force models \cite{yamaguchi2011you, yi2015understanding} had been extensively applied for modelling human navigational behaviour. They rely on the attractive and repulsive forces between pedestrians to predict motion. However as shown in \cite{alahi2016social,fernando2017soft+,fernando2018tree} these methods ill represent the structure of human decision making by modelling the behaviour with just a handful of parameters. 

One of the most popular deep learning methods for predicting human behaviour is the Social LSTM model of \cite{alahi2016social}, which removed the need for hand-crafted features by using LSTMs to encode and decode trajectory information. This method is further augmented in \cite{fernando2017soft+} where the authors incorporate the entire trajectory of the pedestrian of interest as well as the neighbouring pedestrians and extract salient information from these embeddings through a combination of soft and hardwired attention. Similar to \cite{fernando2017soft+} the works in \cite{zou2018understanding,bartoli2017context,varshneya2017human} also highlight the importance of fully capturing context information. However these methods all consider short-term temporal context in the given neighbourhood, completely discarding scene structure and the longterm scene context.

\subsection{Neural Memory Architectures}
\label{sec:nma}
External memory modules are used to store historic information, and learn to automatically store and retrieve important facts to aid future predictions. Many approaches across numerous domains \cite{kaiser2015neural, malinowski2014multi,fernando2017going,fernando2018tree,fernando2017Learning,fernando2018task} have utilised memory modules to aid prediction, highlighting the importance of stored knowledge for decision making. However existing memory structures are one dimensional modules which completely ignore the environmental spatial structure. This causes a significant hindrance when modelling human navigation, since they are unable to capture the map-like structures humans use when navigating \cite{madl2016exploring,epstein2017cognitive}. 

The work of Parisotto et al. \cite{parisotto2018} proposes an interesting extension to memory architectures where they structure the memory as a 3D block, preserving spatial relationships. However, when generating memory output they rely on a static convolution kernel to summarise the content, failing to generate dynamic responses and propagate salient information from spatial locations to the trajectory prediction module, where multiple humans can interact in the environment. 

Motivated by the hierarchical sub-map structure humans use to navigate \cite{derdikman2010manifold,gobet2001chunking}, we model our spatiotemporal memory with gated St-LSTM cells, which are arranged hierarchically in a grid structure.

\subsection{Multimodal information fusion}
\label{sec:mif}

Multimodal information fusion addresses the task of integrating inputs from various modalities and has shown superior performance compared to unimodal approaches  \cite{deng2014deep,bhatt2011multimedia} in variety of applications \cite{huang2017learning,yuan2017ffgs,arevalo2017gated}. The simplest approach is to concatenate features to obtain a single vector representation \cite{kiela2014learning,pei2013unsupervised}. However it ignores the relative correlation between the modalities \cite{arevalo2017gated}.

More complex fusion strategies include concatenating higher level representations from individual modalities separately and then combining them together, enabling the model to learn the salient aspects of individual streams. In this direction, attempts were made using Deep Boltzmann Machines \cite{srivastava2012multimodal} and neural network architectures \cite{coates2011importance,kiros2014multimodal}. 

In\cite{fernando2017going} the authors explore the importance of capturing both short and longterm temporal context when performing feature fusion, utilising separate neural memory units for individual feature streams and aggregating the temporal representation during fusion. Yet this fails to preserve the spatial structure, restricting its applicability when modelling human navigation.

\section{Architecture}

In this section we introduce the encoding scheme utilised to embed the trajectory information of the pedestrian of interest and their neighbours; the structure and the operations of the proposed hierarchical memory; how to utilise memory output to enhance the future trajectory prediction; and an architecture for effectively coupling multimodal information streams through structured memories. 

\subsection{Embedding local neighbourhood context}
\label{sec:local_neighbourhood}
In order to embed the current short-term context of the pedestrian of interest and the local neighbourhood, we utilise the trajectory prediction framework proposed in \cite{fernando2017soft+}.  Let the observed trajectory of pedestrian $k$ from frame 1 to frame $T_{obs}$ be given by,

\begin{equation}
X^k=[(x_1,y_1), (x_2,y_2), \ldots, (x_{T_{obs}},y_{T_{obs}})] ,
\end{equation}

where the trajectory is composed of points in a 2D Cartesian grid. Similar to \cite{fernando2017soft+} we utilise the soft attention operation \cite{chorowski2015attention} to embed the trajectory information from the pedestrian of interest ($k$) and generate a vector embedding $C^{s,k}_t$. To embed neighbouring trajectories the authors in \cite{fernando2017soft+,our_wacv2} have shown that distance based hardwired attention is efficient and effective. We denote the hardwired context vector as $C_{t}^{h,k}$.

Now we define the combined context vector, $C_{t}^{*,k}$, representing the short-term context of the local neighbourhood of the $k^{th}$ pedestrian as,

\begin{equation}
C_{t}^{*,k}=\mathrm{tanh}([C_{t}^{s,k},C_{t}^{h,k}]),
\label{eq:c_star}
\end{equation}

where $[.,.]$ denotes a concatenation operation. Please see \cite{fernando2017soft+, our_wacv2} for details.

\subsection{Structured Memory Network (SMN)}
Let the structured memory, $M$, be a $l \times W \times H$ block where $l$ is the embedding dimension of $p_t^k$. $W$ is the vertical extent of the map and $H$ is the horizontal extent. We define a function $\psi(x,y)$ which maps spatial coordinates $(x,y)$ with $x \in \mathbb{R}$ and $y \in \mathbb{R}$ to a map grid $(x', y')$ where $x' \in {0, \ldots, W}$ and $y' \in {0, \ldots, H}$. The works of \cite{fernando2017soft+,our_wacv2} have shown that the context embeddings $C_{t}^{*,k}$ capture the short-term context of the pedestrian of interest and the local neighbourhood. Hence we store these embeddings in our structured memory as it represents the temporal context in that grid cell. The operations of the proposed structured memory network (SMN) can be summarised as follows,

\begin{equation}
h_t= \mathrm{read}({M_t}),
\end{equation}
\begin{equation}
\beta^{(x', y')}_{t+1}= \mathrm{{write}(C_{t}^{*,k}, M_t^{(x',y')})},
\end{equation}
\begin{equation}
M_{t+1}= \mathrm{{update}}(M_t, w^{(x', y')}_{t+1}).
\end{equation}

The following subsections explain these three operations. 

\subsubsection{Hierarchical Read Operation}

\begin{figure}[t]
\begin{center}
   \includegraphics[width=0.8\linewidth]{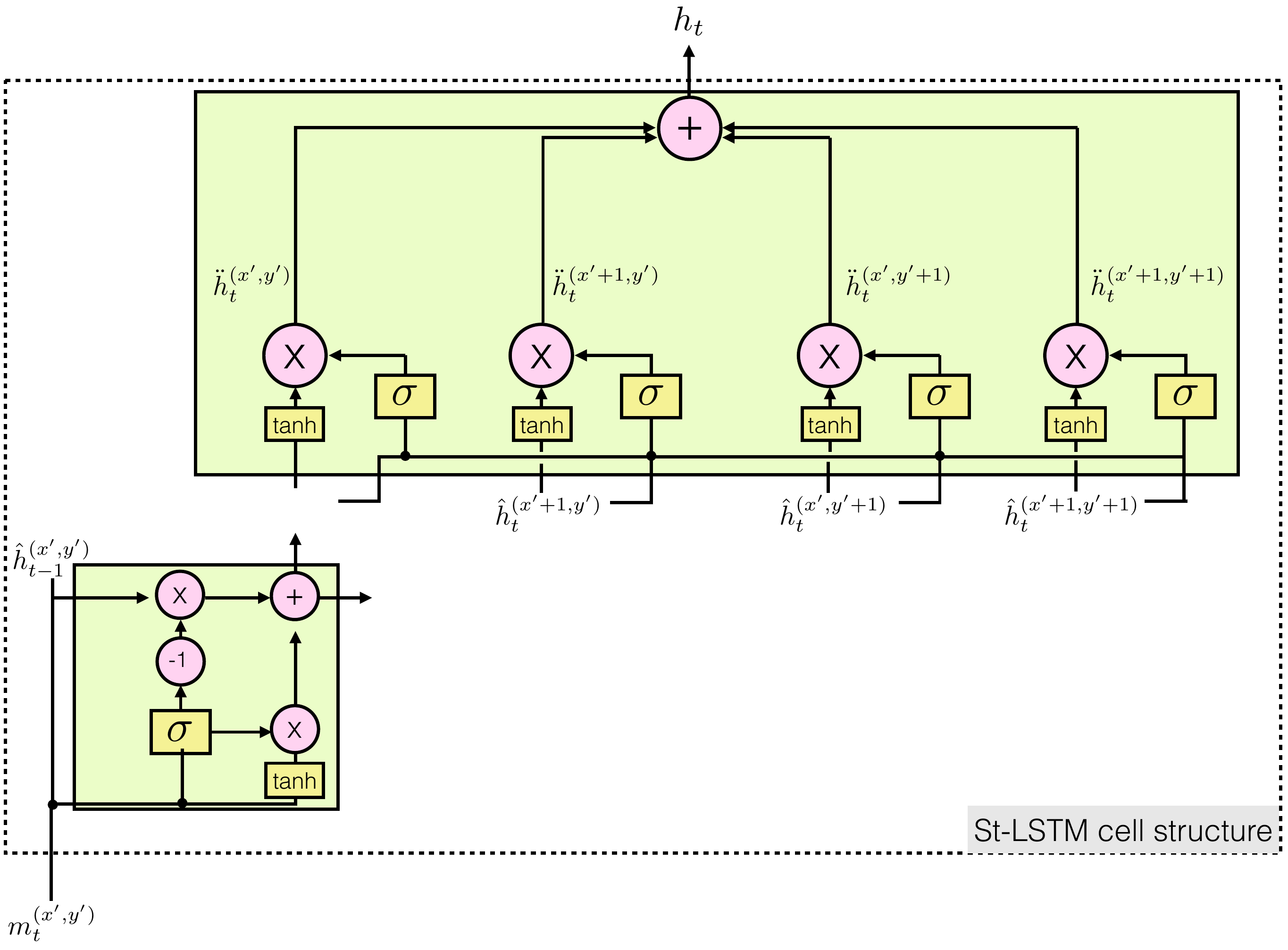}
\end{center}
   \caption{The operations of the proposed St-LSTM cell. It considers the current representation of the respective memory cell and the 3 adjacent neighbours as well as the previous time step outputs and utilises gated operations to render the output in the present time step.}
\label{fig:gates}
\end{figure}

The read operation outputs output a vector, $h_t$, capturing the most salient information from the entire memory for decision making in the present state. We define a hierarchical read operation which passes the current memory representation, $M_t$, through a series of gated, structured LSTM (St-LSTM) cells arranged in a grid like structure. Fig. \ref{fig:gates} depicts the operations of the proposed St-LSTM cells. 

\begin{figure*}
\begin{center}
\includegraphics[width=0.7\linewidth]{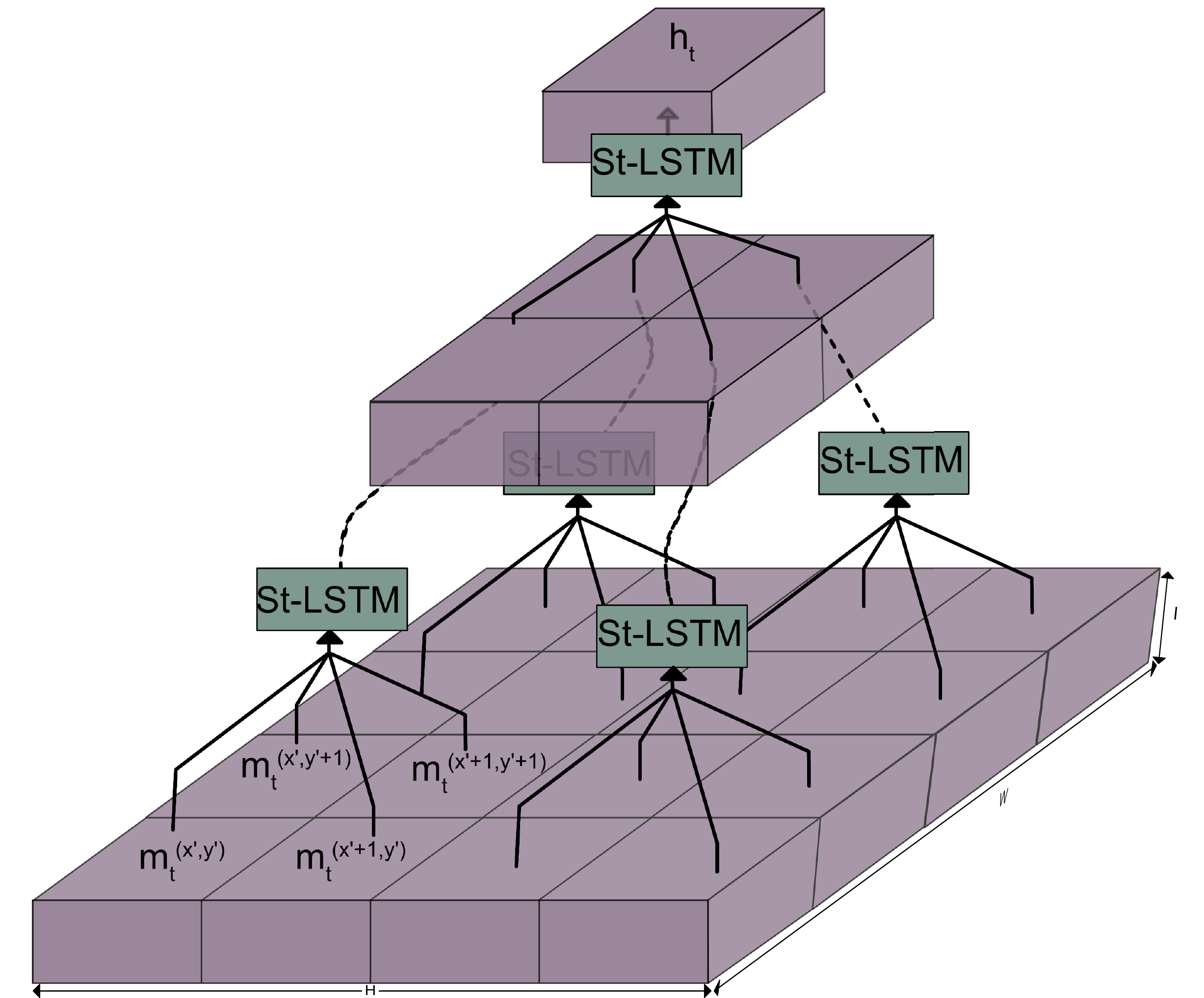}
\end{center}
   \caption{Utilisation of proposed St-LSTM cell to generate a hierarchical embedding of the structured memory. In each layer we summarise the content of 4 adjacent neighbours via propagating the most salient information to the layer above. The process is repeated until we generate a single vector representation of the entire memory block.}
\label{fig:memory}
\end{figure*}

Let the content of $(x', y')$ memory cell at time $t$ be represented by $m_t^{(x', y')}$ and the three adjacent cells be represented by $m_t^{(x'+1, y')}, m_t^{(x', y'+1)} $and $m_t^{(x'+!, y'+1)} $. As shown in Fig. \ref{fig:memory}, we first pass the current state of the memory cell through an input gate to decide how much information to pass through the gate and how much information to gather from the previous hidden state of the that particular cell, $\hat{h}_{t-1}^{(x', y')}$. This operation is given by,

%

\begin{align}
\begin{split}
z_{t}^{(x', y')}= \sigma(w_z^{(x', y')}[m_t^{(x', y')}, \hat{h}_{t-1}^{(x', y')}]) ,
\\
\hat{o}_{t}^{(x', y')}= \mathrm{tanh}([m_t^{(x', y')}, \hat{h}_{t-1}^{(x', y')}]) .
\end{split}
\end{align}

Then we generate the new hidden state of the cell using,
\begin{equation}
\hat{h}_{t}^{(x', y')}= z_t^{(x', y')}\hat{o}_t^{(x', y')} +(1-z_t^{(x', y')})\hat{h}_{t-1}^{(x', y')},
\end{equation}

and pass the hidden state of that particular cell as well as the hidden states of the adjacent cell through a composition gate function which determines the amount of information to be gathered from each of the cells as, 
\begin{equation}
q_{t}^{(x', y')}= \sigma(w_q^{(x', y')}[\hat{h}_t^{(x', y')}, \hat{h}_{t}^{(x'+1, y')}, \hat{h}_{t}^{(x', y'+1)}, \hat{h}_{t}^{(x'+1, y'+1)}]).
\end{equation}

Now we can generate the augmented state of the cell $(x', y')$ as, 
\begin{equation}
\ddot{h}_{t}^{(x', y')}= \mathrm{tanh}(\hat{h}_t^{(x', y')})q_t^{(x', y')}.
\end{equation}

We perform the above operations to the rest of the group of 3 cells: $(x'+1, y'), (x', y'+1)$ and $(x'+1, y'+1)$; and generate the representations $\ddot{h}_{t}^{(x'+1, y')}, \ddot{h}_{t}^{(x', y'+1)}$ and $\ddot{h}_{t}^{(x'+1, y'+1)}$ respectively. Then the feature embedding representation, $h_t$, of the merged 4 cells in the next layer of the memory is given by,

\begin{equation}
h_{t}= \ddot{h}_t^{(x', y')} + \ddot{h}_t^{(x'+1, y')}+  \ddot{h}_t^{(x', y'+1)} +  \ddot{h}_t^{(x'+1, y'+1)} .
\label{eq:h_t}
\end{equation}

We repeat this process for all cells $(x', y')$ where $x' \in {0, \ldots, W}$ and $y' \in {0, \ldots, H}$. Note that as we are merging four adjacent cells we have $W/2$ and $H/2$ St-LSTM cells in the immediate next layer of the memory block. We continue merging cells until we are left with one cell summarising the entire memory block. We denote the hidden state of this cell as $h_{t}$.

\subsubsection{Write Operation}

Given the current position of the pedestrian of interest at $(x, y)$, we first evaluate the associated location in the map grid by passing $(x, y)$ through function, $\psi$, such that,

\begin{equation}
(x', y')= \psi(x, y).
\end{equation}

Then we retrieve the current memory state of $(x', y')$ as, 
\begin{equation}
m_t^{(x', y')}= M_t^{(x', y')}.
\end{equation}

Then by utilising the above stated vector and the short-term context of the pedestrian of interest, $C^{*}_t$, we define a write function which generates a write vector for memory update, 

\begin{equation}
\beta_{t+1}^{(x', y')}= \mathrm{LSTM}_w(c^{*}_t, m_t^{(x', y')}).
\end{equation}

\subsubsection{Update Operation}

We update the memory map for the next time step by,

\begin{equation}
M^{(a, b)}_{t+1} = \begin{cases} \beta^{(x', y')}_{t+1} &\mbox{for  } (a, b) =  (x', y')\\ 
M^{(a, b)}_t &\mbox{for  } (a, b) \neq  (x', y') \end{cases}
\end{equation}

The new memory map is equal to to the memory map at the previous time step except for the current location of the pedestrian where we completely update the content with the generated write vector.

\subsection{Trajectory prediction with structured memory hierarchies}

We utilise the combined context vector $C_{t}^{*,k}$ representing the short-term context of the local neighbourhood of the $k^{th}$ pedestrian, and the generated memory output $h_t$ to generate an augmented vector for the context representation,
 
 \begin{equation}
\bar{c}_{t}^{(k)}= \mathrm{tanh}([C^{*,k}_t, h_t]),
\label{eq:final_combined}
\end{equation}
which is used to predict the future trajectory of the pedestrian $k$,

 \begin{equation}
Y_t= \mathrm{LSTM}(p_{t-1}^k, \bar{c}_{t}^{(k)}, Y_{t-1}).
\label{eq:track_pred}
\end{equation}

\subsection{Coupling multimodal information to improve prediction}
\label{sec:coupling_multi_modal}

Using multimodal input streams allows us to capture different semantics that are present in the same scene, and compliment individual streams. For instance, in a surveillance setting, radar and video fusion is widely utilised \cite{bostrom2017reducing,roy2011automated} as radar offers better coverage in the absence of visual light, however has a lower frame rate ($<$5 fps) compared to video ($\sim$25fps) which records more fine grained motion. 

As pointed out in \cite{fernando2017going}, simply concatenating data from both streams leads to information loss as they contain information at different granularities. Hence it is vital to jointly back propagate among the information streams to learn the important aspects of each. Therefore, we capture streams through separate memory modules, and perform gated coupling of memory hierarchies. 

We denote the two synchronised input modalities as $I$ and $R$, where the trajectory of pedestrian $k$ observed in stream $I$ is denoted as $X^{k}_{I}$ and the same pedestrian trajectory observed in stream $R$ is given as $X^{k}_{R}$. We pass each stream separately through the local neighbourhood embedding mechanism proposed in Sec. \ref{sec:local_neighbourhood} and generate vector embeddings $C^{*,k}_{t,I}$ and $C^{*,k}_{t,R}$ respectively. We embed these through individual memory blocks denoted as $M_{t,I}$ and $M_{t,R}$. In the absences of such trajectories (i.e due to poor coverage, occlusions, \ldots), we evaluate only the neighbourhood embeddings $C^{h,k}_{t,R}$ and use them as $C^{*,k}_{t,R}$. 

\begin{figure}[t]
\begin{center}
   \includegraphics[width=0.7\linewidth]{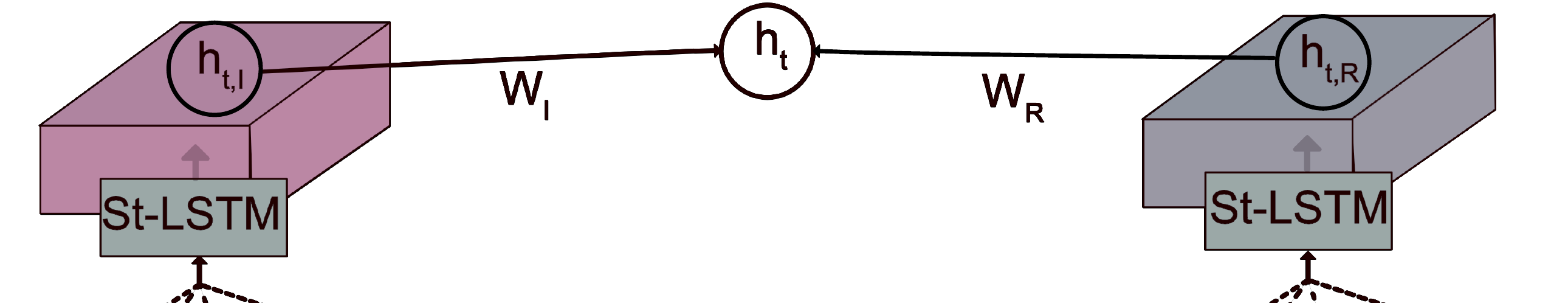}
\end{center}
   \caption{Coupling multimodal information through multiple memory modules. The information from each modality is stored separately. Note that the figure shows only the top most layer in each memory.}
   \vspace{-5mm}
\label{fig:fusion}
\end{figure}

After the hierarchical gated operations, let the memory output generated using Eq. \ref{eq:h_t} from memory $M_{t,I}$ at time instance $t$ be denoted by $h_{t,I}$ and memory $M_{t,R}$ be denoted as $h_{t,R}$. For simplicity, in Fig. \ref{fig:fusion} we consider 2 input streams, however the proposed coupling mechanism is flexible and is able to handle any number of modalities. 

Motivated by \cite{arevalo2017gated,kiela2018efficient} we perform gated modality fusion such that,

%


\begin{align}
\label{eq:multi_first}
\begin{split}
 \bar{h}_{t,I}=\mathrm{tanh}(W_Ih_{t,I}) ,
\\
\bar{h}_{t,R}=\mathrm{tanh}(W_Rh_{t,R}) ,
\\
\nu=\sigma(W_{\nu}[\bar{h}_{t,I},\bar{h}_{t,R}]),
\end{split}
\end{align}

where $W_I$ and $W_R$ are the weights for the respective memories and $W_{\nu}$ is the weight of the fusion gate. This can be seen as performing attention from one modality over the other where each modality determines the amount of information to flow from the other. We then obtain the combined feature vector,

\begin{equation}
h_{t}=\nu\bar{h}_{t,I} + (1- \nu)\bar{h}_{t,R} ,
\end{equation}

and augment Eq. \ref{eq:final_combined} to utilise information from both streams,

 \begin{equation}
\bar{C}_{t}^{(k)}= \mathrm{tanh}([C^{*,k}_{t,I},C^{*,k}_{t,R}, h_t]),
\label{eq:multi_final}
\end{equation}

and predict the future trajectory using Eq. \ref{eq:track_pred}. In contrast to \cite{bostrom2017reducing,roy2011automated} where simple concatenation of multimodal data is used, the proposed multi-memory architecture allows the model to store salient information of individual streams separately and propagate it effectively to the decision making process. We denote this model as $SMN(I+R)$ as it couples $I$ and $R$ streams to the $SMN$ model.

\section{Evaluation and Discussion}

\subsection{Datasets}
We present the experimental results for the single modal framework on the publicly available New York Grand Central (GC) \cite{yi2015understanding} dataset. The Grand Central dataset consist of 12,600 trajectories. For training, testing and validation we use the same splits defined in \cite{fernando2017soft+}. Due to the unavailability of public multimodal pedestrian trajectory data, we introduce a new large scale dataset \footnote{available at https://github.com/qutsaivt/SAIVTMultiSpectralTrajectoryDataset}. Pedestrian trajectories from a CCTV surveillance feed (I) and Radar (R) streams, for 32 hours, were collected and synchronised. Please refer to the supplementary material for statistics, calibration and synchronisation details of the dataset.

\subsection{Evaluation Metrics}
\label{sec:eval_metrics}
Following \cite{alahi2016social,fernando2017soft+} we evaluate the performance with the following 3 error metrics: Average displacement error (ADE), Final displacement error (FDE) and Average non-linear displacement error (n-ADE). Please refer to \cite{alahi2016social,fernando2017soft+} for details. 

\subsection{Evaluation of trajectory prediction with single modal data}
\label{sec:eval_single_modal}
The evaluation of single modal trajectories is conducted on the GC dataset \cite{yi2015understanding}. We compare our model against 6 state of the art baselines. The first baseline is the Social Force (SF) model of \cite{yamaguchi2011you}. It requires the destination of the pedestrian as input, and a linear SVM is trained with ground truth destination areas for this task. The next baseline is the Social LSTM (So-LSTM) model of \cite{alahi2016social}. It requires the neighbourhood size as a hyper-parameter and is set to 32px. The soft $+$ hardwired attention model from \cite{fernando2017soft+} (SHA) does not posses any memory and computes the trajectory prediction by modelling the local neighbourhood of the pedestrian of interest. We also consider the Tree Memory Network (TMN) \cite{fernando2018tree} which models the memory as a tree structure. This model uses the hyper parameter $\delta$, which defines the length of the memory as it structures a flat memory vector as a tree. We also evaluate the Neural Map (NM) model introduced in \cite{parisotto2018}. The pedestrian of interest's trajectory is embedded using a soft attention mechanism as defined in Sec. \ref{sec:local_neighbourhood} and is stored in the memory. To provide a fair comparison, we also augment the NM module with the neighbourhood embeddings, $C^{s,k}_{t}$ and $C^{h,k}_{t}$, combine these with the memory output vector generated from the NM as in Eq. \ref{eq:final_combined}. We define this model as NMA.

To provide a direct comparison among baselines we set the hidden state dimensions of So-LSTM, SHA, TMN, NM, NMA and the proposed SMN model to be 30 units. As the models NM, NMA and SMN have a map width (W) and map height (H) as hyper-parameters, we evaluate different memory sizes. Similarly, for TMN we evaluate different memory lengths $\delta$. Please refer to the supplementary material for those evaluations. Best results are shown in Tab. \ref{tab:gc_eval}. To evaluate the relative performance of each model, we observe the trajectory for 20 frames and predict the future trajectory for the next 20 frames.

\begin{table*}[htbp]
\centering
\begin{tabular}{|C{3.5cm}|C{2cm}|C{2cm}|C{2cm}|}
\hline
\multirow{2}{*}{Method} & \multicolumn{3}{c|}{Metric} \\ \cline{2-4} 
                        & ADE     & FDE    & n-ADE    \\ \hline
SF \cite{yamaguchi2011you}                     &3.364         &5.808        & 3.983         \\ \hline
So-LSTM \cite{alahi2016social}                 &1.990         &4.519        & 1.781    \\ \hline
SHA \cite{fernando2017soft+}                    & 1.096        & 3.011       & 0.985          \\ \hline
TMN ($\delta$=64) \cite{fernando2018tree}             & 2.982         &  4.989       &  2.780        \\ \hline
NM (W=H=64) \cite{parisotto2018}             & 2.505        & 4.151       &   2.432       \\ \hline
NMA (W=H=64)            & 1.466       &  3.811      &  1.445        \\ \hline 
SMN (W=H=128)            &  \textbf{0.891}       & \textbf{2.899}       & \textbf{0.814}         \\ \hline

\end{tabular}
\caption{Quantitative results with the GC dataset \cite{yi2015understanding} for Social Force (SF) \cite{yamaguchi2011you}, Social LSTM (So-LSTM) \cite{alahi2016social}, Soft $+$ Hardwired Attention (SHA) \cite{fernando2017soft+}, Tree Memory Network (TMN) \cite{fernando2018tree}, Neural Map (NM) \cite{parisotto2018}, Neural Map Augmented (NMA) and the proposed Structured Memory Network (SMN) models. In all the methods forecast trajectories are of length 20 frames. The measured error metrics are as in Sec. \ref{sec:eval_metrics}.}

\label{tab:gc_eval}
\end{table*}

From the results tabulated in Tab. \ref{tab:gc_eval} we observe poor performance in the SF model due to its lack of capacity to model long-term history. Models So-LSTM and SHA utilise short-term history from the pedestrian of interest and the local neighbourhood and generate improved predictions accordingly. 

The lack of spatial structure and context modelling in the TMN module leads to it's poor performance despite it's long-term history modelling capacity. Comparing the NM and NMA models, the performance increase from NM and NMA is due to the addition of local context, highlighting the importance of capturing both long and short-term context. The NMA model attains improved performance due to the improved modelling of the local neighbourhood, and the structured memory; however when compared to the SHA model it fails to propagate salient spatiotemporal information from the structured memory to aid the decision making. This is due to the static kernel used when generating the memory output. In contrast, we map the memory output hierarchically using the proposed St-LSTM cells and propagate salient information to the upper layer, enabling efficient information transfer to the prediction model. The proposed gated architecture considers the evolution of memory over time, where multiple humans can interact with the environment, changing the state of multiple spatial locations. Hence we are able to generate dynamic responses instead of passing the information through a static convolution kernel as in NM and NMA; enabling superior performance even with large memory sizes.

\begin{figure*}[htbp]
\begin{center}
\subfigure[]{\includegraphics[width = .25 \textwidth]{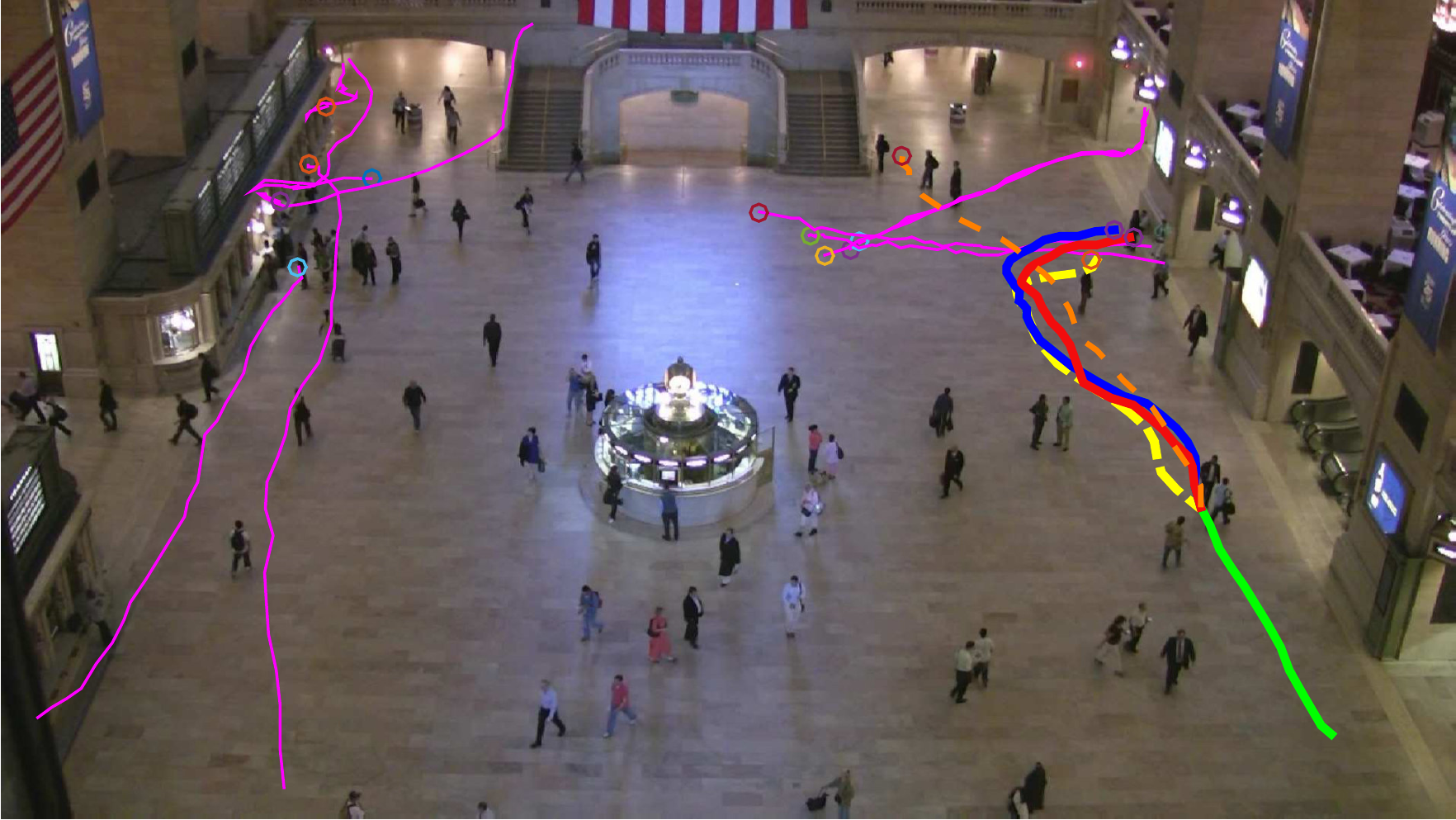}}  
\subfigure[]{\includegraphics[width = .25 \textwidth]{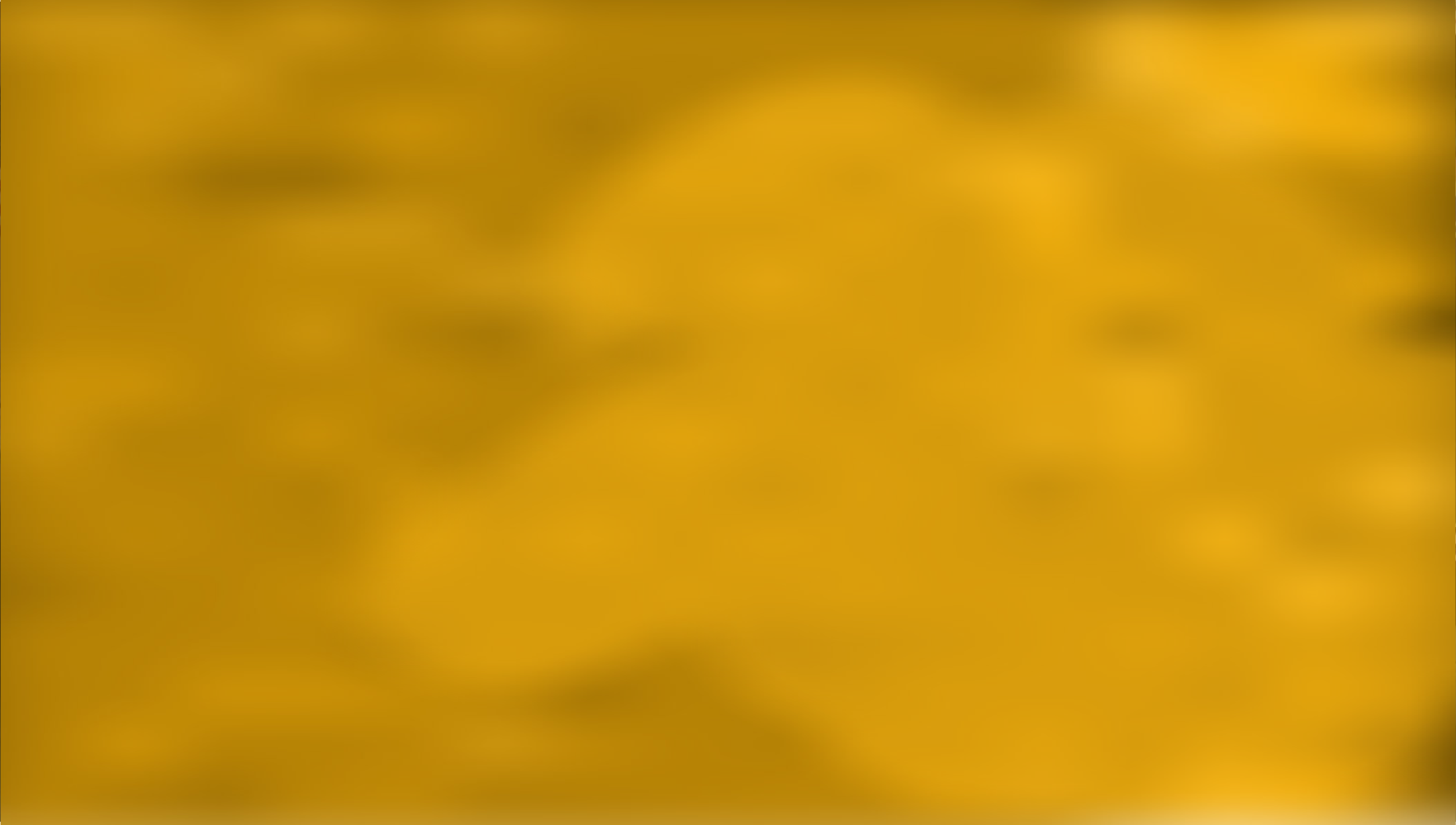}}  
\subfigure[]{\includegraphics[width = .25 \textwidth]{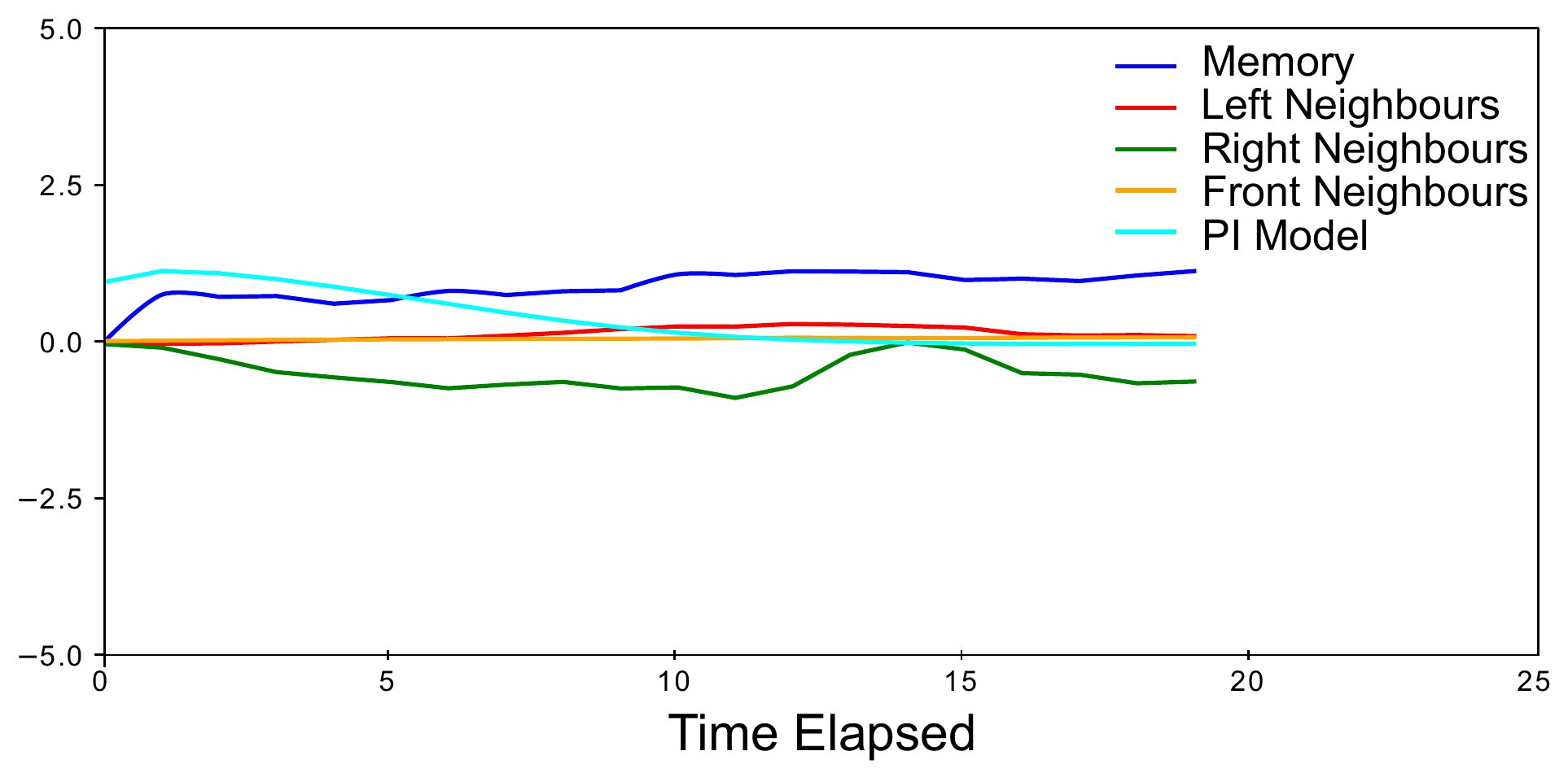}}  
\subfigure[]{\includegraphics[width = .25 \textwidth]{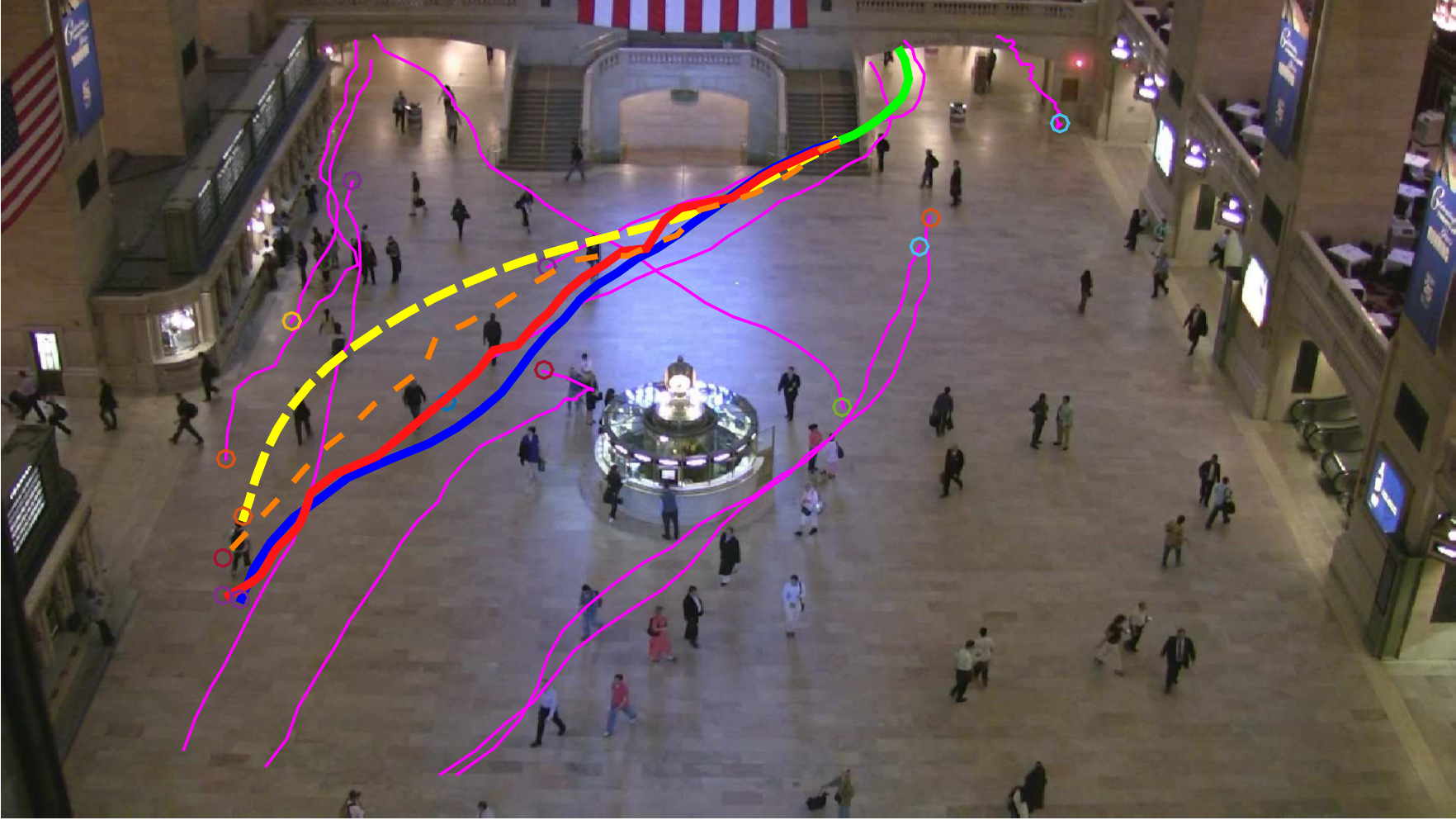}}  
\subfigure[]{\includegraphics[width = .25 \textwidth]{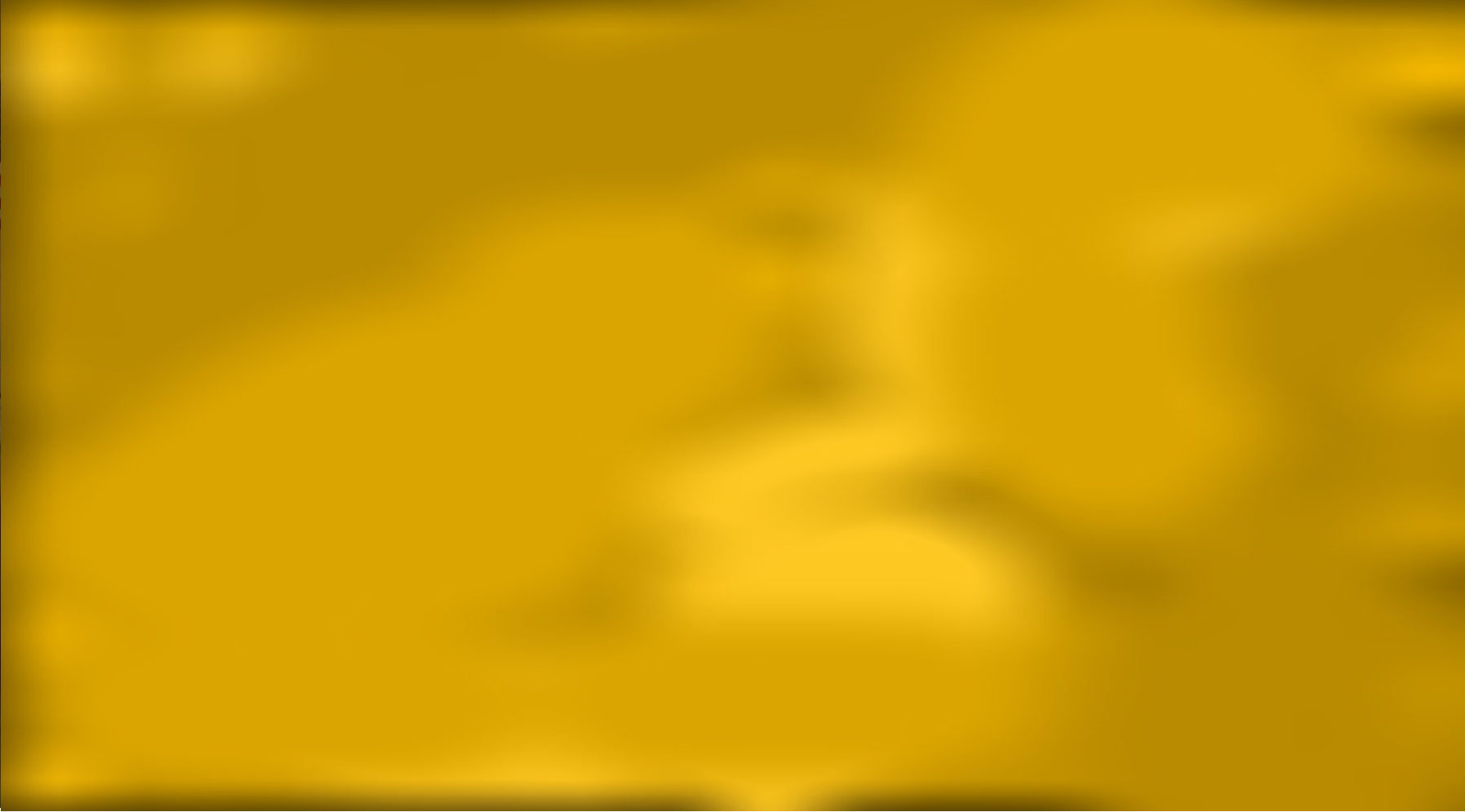}}  
\subfigure[]{\includegraphics[width = .25 \textwidth]{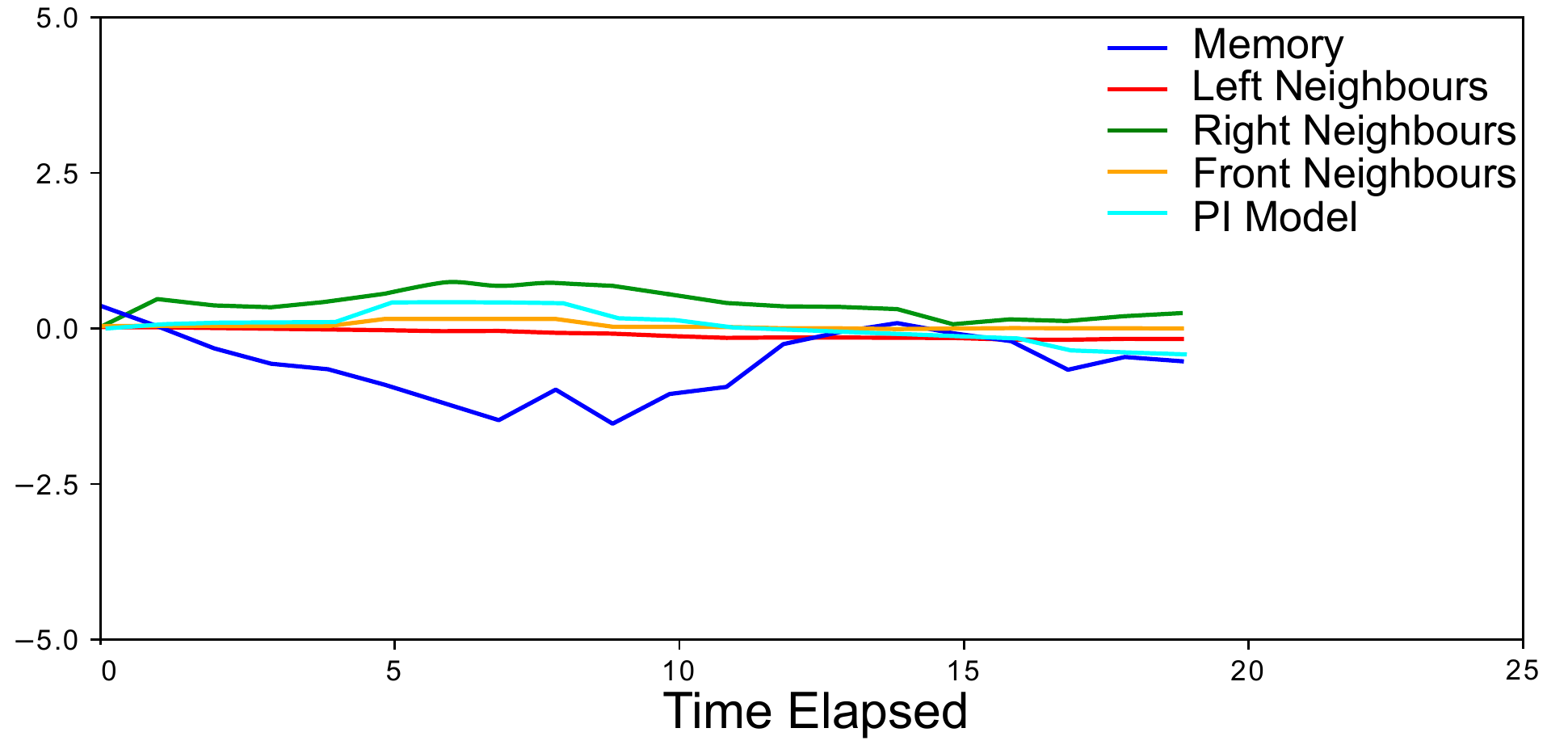}}  
\end{center}
\caption{Qualitative results for the GC dataset \cite{yi2015understanding}: Given (in green), Ground Truth (in blue), Neighbouring (in purple) and Predicted trajectories from SMN model (in red), from SHA model (in yellow), from NMA model (in orange) along with the respective structured memory activations and relative activation contribution of each component in the prediction module. Please note that in the structured memory activations the intensity of the colour represents the degree of the activation and has been manually aligned with the figure in the first column for the clarity of visualisation.}
\label{fig:fig_qualitative}
\end{figure*}

We present a qualitative evaluation of the proposed SMN model with the SHA and NMA baselines in Fig. \ref{fig:fig_qualitative}. We selected these baselines as they provide the highest comparative results. The trajectories are shown in the first column where the observed part of the trajectory is denoted in green, the ground truth observations in blue, neighbouring trajectories are in purple and the predicted trajectories are shown in red (SMN), yellow (SHA) and orange (NMA).

When observing the qualitative results it can be clearly seen that the proposed SMN model generates better trajectory predictions compared to the state-of-the-arts. For instance in Fig. \ref{fig:fig_qualitative} (a) and (d) we observe significant deviation of the predictions of SHA and NMA models from the ground truth. However the proposed SMN model has been able to anticipate the pedestrian motion more accurately with the improved context modelling. 

From the memory activation visualisations, it is evident more attention is given to cells surrounding the trajectory of the pedestrian of interest and the neighbours. Varying levels of attention are given to the cells occupied by the neighbours. However by passing this information through the proposed gated St-LSTM cells the proposed model is able to learn salient information among the passed activations from the layer below. This can be verified by observing the relative activation plots presented in the 3rd column of Fig. \ref{fig:fig_qualitative}. While in general more attention is given to the encoded trajectory information from the pedestrian of interest (PI model), in cases such as Fig. \ref{fig:fig_qualitative} (c) more attention is given to the historic neighbourhood embeddings present in memory, where as in Fig. \ref{fig:fig_qualitative} (f) the model gives more attention to the neighbours. This verifies our hypothesis that both current context information encoded within the motion of pedestrian of interest and the neighbouring trajectories as well as the information from the long-term history that preserves the structural integrity, is vital for prediction. Refer to the supplementary material for more qualitative evaluations.

\subsection{Evaluation of trajectory prediction with multimodal data}
\label{sec:eval_multi_modal}
The evaluation of  multi-modal trajectories is conducted on the proposed multi-modal dataset. We compared our proposed model, SMN(I$+$R), with 4 state of the art baselines. In the first baseline, SHA(I$+$R), we concatenate the embeddings $C^{*,k}_{I,t}$ and $C^{*,k}_{R,t}$ for the $I$ and $R$ modalities directly to generate the augmented vector representation, $\bar{C}_{t}^{(k)}$, and use it in Eq. \ref{eq:track_pred} to generate the prediction. The work of Fernando et al. \cite{fernando2017going} introduces a multi-modal extension to the TMN module. We use this model, TMN(I$+$R), as our next baseline. We extend the NM and NMA architectures (see Sec. \ref{sec:eval_single_modal}) to handle multi-modal data. Similar to TMN(I$+$R) model we use multiple memories to store each input streams and pass the memory outputs through Eq. \ref{eq:multi_first} to generate predictions. The augmented models are denoted NM(I$+$R) and NMA(I$+$R) in the evaluations. For models NM(I$+$R), NMA(I$+$R) and SMN(I$+$R) we set the map width (W) and map height (H) to be 128 and for TMN(I$+$R) we set the memory length $\delta$=64, as this provided the best accuracies in Sec \ref{sec:eval_single_modal}.

Following the previous experiment, we observe the trajectory for 20 frames and predict the trajectory for the next 20 frames. After filtering out short and fragmented trajectories we are left with 40,800 trajectories. We randomly selected 28,560 trajectories for training, 10,200 for testing and 2,040 for validation. 

\begin{table*}[htbp]
\centering
\begin{tabular}{|C{3.5cm}|C{2cm}|C{2cm}|C{2cm}|}
\hline
\multirow{2}{*}{Method} & \multicolumn{3}{c|}{Metric} \\ \cline{2-4} 
                        & ADE     & FDE    & n-ADE    \\ \hline
SHA(I$+$R) \cite{fernando2017soft+}                    & 1.245        & 1.654       & 1.454        \\ \hline
TMN(I$+$R) \cite{fernando2017going}               & 2.901        & 3.169       &3.001     \\ \hline
NM(I$+$R)  \cite{parisotto2018}                  & 2.015        & 2.741       &2.344           \\ \hline
NMA(I$+$R)             &1.325          &1.814         &1.558          \\ \hline \hline
SMN(I$+$R)            & \textbf{0.979}       &  \textbf{0.998}      & \textbf{1.036}         \\ \hline 
\end{tabular}
\caption{Quantitative results with the proposed multimodal dataset for, Soft $+$ Hardwired Attention (SHA(I$+$R)) \cite{fernando2017soft+}, Tree Memory Network (TMN(I$+$R)) \cite{fernando2018tree}, Neural Map (NM(I$+$R)) \cite{parisotto2018}, Neural Map Augmented (NMA(I$+$R)) and the proposed Structured Memory Network (SMN(I$+$R)) models. In all the methods forecast trajectories are of length 20 frames. Error metrics are defined in Sec. \ref{sec:eval_metrics}.}
\label{tab:multimodal_eval}
\end{table*}

Similar to the evaluations in Sec. \ref{sec:eval_single_modal}, we observe poor performance from TMN(I$+$R) and NM(I$+$R) due to their inability to capture local neighbourhood information. However we observe a significant reduction in the performance gap between SHA(I$+$R) and NMA(I$+$R), compared to the that in Tab.\ref{tab:gc_eval}, which is a result of the naive fusion method used in the former model. SHA(I$+$R) simply concatenates the two modes together, and as such the model lacks the capacity to capture salient information from individual modes. In contrast, by capturing long-term temporal dependencies of the two modalities, the memory based coupling mechanism yields better predictions. We further augment this process in SMN(I$+$R) by utilising the St-LSTM cells to hierarchically capture salient information from each mode. This enables the model to jointly back propagate through the two modalities and learn the strengths and weaknesses of each, effectively complimenting the prediction module with the additional information stream. Please refer to supplementary material for qualitative evaluations of the proposed SMN(I$+$R) model with the SHA(I$+$R)  and NMA(I$+$R)  baselines.
%

\subsection{Ablation Experiments}
To further demonstrate the effectiveness of our proposed fusion approach, we conduct a series of ablation experiments, identifying the crucial components of the proposed architecture. In the same settings as the experiment in Sec. \ref{sec:eval_multi_modal}, we compare the SMN(I$+$R) (proposed) method to a series of counterparts constructed by removing components of the model as follows:
\begin{itemize} 
\item \textbf{SA(I)}: Uses only the soft attention context vector, $C^{s,k}_{t}$, and data from the image stream (I) for trajectory prediction. 
\item \textbf{SHA(I)}: Uses both soft ($C^{s,k}_{t}$) and hardwired ($C^{h,k}_{t}$) attention vectors and data from image stream (I) for trajectory prediction
\item \textbf{SMN(I)}: Uses the proposed SMN model and data from Image (I) stream.
\item \textbf{SA(R)}: Similar to SA-I but uses data from the Radar (R) stream.
\item \textbf{SHA(R)}: Similar to SHA-I but uses data from the Radar (R) stream.
\item \textbf{SMN(R)}: Similar to SMN-I but uses data from the Radar (R) stream.
\item \textbf{SA(I$+$R)}: SA model that directly concatenates $C^{s,k}_{t,I}$ and $C^{s,k}_{t,R}$ and generates a vector embedding for Eq. \ref{eq:track_pred}.
\item \textbf{SHA(I$+$R)}: SHA model that directly concatenates $C^{*,k}_{t,I}$ and $C^{*,k}_{t,R}$ and generates a vector embedding for Eq. \ref{eq:track_pred}.
\item \textbf{SMN(I$+$R)}: Uses the model proposed in Sec. \ref{sec:coupling_multi_modal}.
\end{itemize} 

Note that for all $SMN$ models we used $W=H=128$.

\begin{table*}[htbp]
\centering
\begin{tabular}{|C{3.5cm}|C{2cm}|C{2cm}|C{2cm}|}
\hline
\multirow{2}{*}{Method} & \multicolumn{3}{c|}{Metric} \\ \cline{2-4} 
                        & ADE     & FDE    & n-ADE    \\ \hline
SA(I)                    & 2.012        &3.011        &2.190          \\ \hline
SHA(I)               &1.235         &2.731        &1.442     \\ \hline
SMN(I)                    &1.029         & 1.104       &  1.092         \\ \hline
SA(R)             & 2.259         & 3.312        &2.261          \\ \hline
SHA(R)            & 1.613        &3.070         & 1.892         \\ \hline
SMN(R)              & 1.198         & 1.330       & 1.288         \\ \hline 
SA(I$+$R)             &1.334         &1.813        &1.579          \\ \hline
SHA(I$+$R)            &1.245        &1.654        & 1.454         \\ \hline \hline
SMN(I$+$R)            & \textbf{0.979}       &  \textbf{0.998}      & \textbf{1.036}         \\ \hline 
\end{tabular}
\caption{Ablation experiment evaluations}
\label{tab:ablation_eval}
\end{table*}

The results of our ablation study are presented in Tab. \ref{tab:ablation_eval}. Models SA(I) and SA(R) perform poorly due to their inability to oversee the neighbourhood context. We observe improved performance in SHA(I) and SHA(R) with the introduction of information from neighbouring pedestrians. The combined information from both modalities contributes to the performance gain we observe in SHA(I$+$R) over the unimodal counterparts, verifying the observations in \cite{deng2014deep,bhatt2011multimedia}. 

Comparing the unimodal SMN(I) and SMN(R) models with the multimodal SHA(I$+$R) model, the former outperforms the latter by a significant margin, emphasising the importance of capturing long-term spatial context, and propagating the information effectively to the prediction model. The introduction of a secondary modality in SMN(I$+$R) further improves the prediction accuracy. 

We would like to further compare the results obtained from the individual models in the $I$ and $R$ streams. We observe a performance boost in modality $I$, due to the finer granularity present in the CCTV stream due to the higher frame rate, compared to the radar stream. Hence extracted trajectories are smoother compared to the trajectories from modality $R$, making it easier to model.  

\subsection{Implementation Details}

We use Keras \cite{chollet2017keras} for our implementation. The SMN and SMN(I$+$R) modules do not require any special hardware (i.e. GPUs) to run. The SMN (W=H=128) model has 152K trainable parameters, and SMN(I$+$R) (W=H=128) has 358K. We ran the test set in Sec. \ref{sec:eval_single_modal} on a single core of an Intel Xeon E5-2680 2.50GHz CPU and the SMN algorithm was able to generate 1000 predicted trajectories with 40, 2 dimensional data points (i.e. using 20 observations to predict the next 20 data points) in 2.791 seconds. In a similar experiment with the test set in Sec. \ref{sec:eval_multi_modal} we were able to generate 1000 predicted trajectories in 11.722 seconds.

\section{Conclusions}
In this paper we propose a method to anticipate complex human motion by analysing structural and temporal accordance. We extend the standard pedestrian trajectory prediction framework by introducing a novel model, Structured Memory Network (SMN), which is able to oversee the long-term history, preserving the structural integrity and improving prediction of pedestrian motion. As an extension to the proposed SMN model, we contribute a novel data driven method to capture salient information from multiple modalities and demonstrate how to incorporate this to enhance prediction. Additionally, we introduce a novel multi-modal pedestrian trajectory dataset, collected from synchronised CCTV and Radar streams, and consisting of 40,000 pedestrian trajectories. Our evaluations on both single and multi-modal datasets demonstrate the capacity of the proposed SMN method to learn complex real world human navigation behaviour.

\subsection*{Acknowledgement}
\small{
This research was supported in part by the Defence Science and Technology (DST) Group under the Defence Science Partnership Program. The authors acknowledge the contribution to the paper by Dr. Jason Williams, Senior Research Scientist, National Security, Intelligence, Surveillance and Reconnaissance Division of DST.
}

%
%
%
\bibliographystyle{splncs04}
\bibliography{egbib}

\end{document}